\documentclass[10pt, a4paper, conference, compsocconf]{IEEEtran}
\ifCLASSINFOpdf
\else
\fi
\usepackage[cmex10]{amsmath}

\usepackage{graphicx}
\newcommand{\comment}[1]{}

\let\oldbibliography\thebibliography
\renewcommand{\thebibliography}[1]{%
  \oldbibliography{#1}%
  \setlength{\itemsep}{0pt}%
}

\begin{document}
%
\title{Toward a Standardized and More Accurate Indonesian Part-of-Speech Tagging}



\author{\IEEEauthorblockN{Kemal Kurniawan}
\IEEEauthorblockA{Kata Research Team\\
Kata.ai\\
Jakarta, Indonesia\\
kemal@kata.ai}
\and
\IEEEauthorblockN{Alham Fikri Aji}
\IEEEauthorblockA{School of Informatics\\
University of Edinburgh\\
Edinburgh, Scotland\\
a.fikri@ed.ac.uk,}
}


%


\IEEEoverridecommandlockouts
\IEEEpubid{\begin{minipage}{\textwidth}
  \vspace{10mm}
  \copyright~2018 IEEE. Personal use of this material is permitted. Permission
  from IEEE must be obtained for all other uses, in any current or future media,
  including reprinting/republishing this material for advertising or
  promotional purposes, creating new collective works, for resale or
  redistribution to servers or lists, or reuse of any copyrighted component or
  this work in other works. The final version of this article is available at
  https://doi.org/10.1109/IALP.2018.8629236.
\end{minipage}}

\maketitle

\begin{abstract}

Previous work in Indonesian part-of-speech (POS) tagging are hard to compare as they are not evaluated on a common dataset. Furthermore, in spite of the success of neural network models for English POS tagging, they are rarely explored for Indonesian.  In this paper, we explored various techniques for Indonesian POS tagging, including rule-based, CRF, and neural network-based models. We evaluated our models on the IDN Tagged Corpus. A new state-of-the-art of 97.47 F1 score is achieved with a recurrent neural network. To provide a standard for future work, we release the dataset split that we used publicly.

\end{abstract}


%
\IEEEpeerreviewmaketitle

\section{Introduction}

Part-of-speech (POS) tagging is a process to tag tokens in a string with their corresponding part-of-speech (e.g., noun, verb, etc). POS tagging is considered as one of the most basic tasks in NLP, as it is usually the first component in an NLP pipeline. This is because POS tags are shown to be useful features in various NLP tasks, such as named entity recognition~\cite{luo2015joint, aguilar2017multi}, machine translation~\cite{sennrich-haddow:2016:WMT, niehues2017exploiting}
and constituency parsing~\cite{dyer2016}. Therefore, for any language, building a successful NLP system usually requires a well-performing POS tagger.

There are quite a number of research on Indonesian POS tagging~\cite{pisceldo2009,wicaksono2010,rashel2014,abka2016}. However, almost all of them are not evaluated on a common dataset. Even when they are, their train-test split are not the same. This lack of a common benchmark dataset makes a fair comparison among these works difficult. Moreover, despite the success of neural network models for English POS tagging~\cite{ma2016,rei2016}, the use of neural networks is generally unexplored for Indonesian. As a result, published results may not reflect the actual state-of-the-art performance of Indonesian POS tagger.

In this work, we explored different neural network architectures for Indonesian POS tagging. We evaluated our experiments on the IDN Tagged Corpus~\cite{dinakaramani2014}. Our best model achieves 97.47 $F_1$ score, a new state-of-the-art result for Indonesian POS tagging on the dataset. We release the dataset split that we used to serve as a benchmark for future work.

\section{Related Work}

Pisceldo et al.~\cite{pisceldo2009} built an Indonesian POS tagger by employing a conditional random field (CRF)~\cite{lafferty2001} and a maximum entropy model. They used contextual unigram and bigram features and achieved accuracy scores of 80-90\% on PANL10N\footnote{http://www.panl10n.net} dataset tagged manually using their proposed tagset. The dataset consists of 15K sentences. Another work used a hidden Markov model enhanced with an affix tree to better handle out-of-vocabulary (OOV) words~\cite{wicaksono2010}. They evaluated their models on the same PANL10N dataset and achieved more than 90\% overall accuracy and roughly 70\% accuracy for the OOV cases. We note that while the datasets are the same, the split could be different. Thus, making a fair comparison between them is difficult.

Dinakaramani et al.~\cite{dinakaramani2014} proposed IDN Tagged Corpus, a new manually annotated POS tagging corpus for Indonesian. The corpus consists of 10K sentences and 250K tokens, and its tagset is different than that of the PANL10N dataset. The corpus is available online.\footnote{https://github.com/famrashel/idn-tagged-corpus} A rule-based tagger is developed in~\cite{rashel2014} using the aformentioned dataset, and is able to achieve an accuracy of 80\%.

One of the neural network-based POS taggers for Indonesian is proposed in~\cite{abka2016}. They used a feedforward neural network with an architecture similar to that proposed in~\cite{collobert2011}. They evaluated their methods on the new POS tagging corpus~\cite{dinakaramani2014} and separated the evaluation of multi- and single-word expressions. They experimented with several word embedding algorithms trained on Indonesian Wikipedia data and reported macro-averaged $F_1$ score of 91 and 73 for the single- and multi-word expression cases respectively. We remark that the choice of macro-averaged $F_1$ score is more suitable than accuracy for POS tagging because of the class imbalance in the dataset. There are too many words with NN as the true POS tag, so accuracy is not the best metric in such case.


\section{Methodology}

\subsection{Dataset}

We used the IDN Tagged Corpus proposed in~\cite{dinakaramani2014}. The corpus contains 10K sentences and 250K tokens that are tagged manually. Due to the small size,\footnote{As a comparison, Penn Treebank corpus for English has 40K sentences.} we used 5-fold cross-validation to split the corpus into training, development, and test sets. We did not split multi-word expressions but treated them as if they are a single token. All 5 folds of the dataset are available publicly\footnote{https://github.com/kmkurn/id-pos-tagging/blob/master/data/dataset.tar.gz} to serve as a benchmark for future work.

\subsection{Baselines}

We used two simple baselines: majority tag (\textsc{Major}) and memorization (\textsc{Memo}). \textsc{Major} simply predicts the majority POS tag found in the training set for all words. \textsc{Memo} remembers the word-tag assignments from the training set and uses them to predict the tags on the test set. If there is an unknown word, it simply outputs the majority tag found in the training set.

\subsection{Comparisons}

\subsubsection{Rule-based tagger}

We adopted a rule-based tagger designed by Rashel et al.~\cite{rashel2014building} as one of our comparisons. Firstly, the tagger tags named entities and multi-word expressions based on a dictionary. Then, it uses MorphInd~\cite{larasati2011indonesian} to tag the rest of the words. Finally, they employ 15 hand-crafted rules to resolve ambiguous tags in the post-processing step. We want to note that we did not use their provided tokenizer since the IDN Tagged Corpus dataset is already tokenized. Their implementation is available online.\footnote{https://github.com/andryluthfi/indonesian-postag}

\subsubsection{Conditional random field (CRF)}

We used CRF~\cite{lafferty2001} as another comparison since it is the most common non-neural model for sequence labeling tasks. We employed contextual words as well as affixes as features. For some context window size $d$, the complete list of features is:
\begin{enumerate}
    \item the current word, as well as $d$ preceding and succeeding words;
    \item two and three leading characters of the current word and $d$ preceding and succeeding words;
    \item two and three trailing characters of the current word and $d$ preceding and succeeding words.
\end{enumerate}
The last two features are meant to capture prefixes and suffixes in Indonesian which usually consist of two or three characters. One advantage of this feature extraction approach is that it does not require language-specific tools such as stemmer or morphological segmenter. This advantage is particularly useful for Indonesian which does not have well-established tools for such purposes. We padded the input sentence with padding tokens to ensure that every token has enough preceding and succeeding words for context window size $d$. For the implementation, we used \texttt{pycrfsuite}.\footnote{https://github.com/scrapinghub/python-crfsuite}

\subsubsection{Neural network-based tagger}

Our neural network-based POS tagger can be divided into 3 steps: embedding, encoding, and prediction. First, the tagger embeds the words and optionally additional features of such words (e.g., affixes). From this embedding process, we get vector representations of the words and the features. Next, the tagger learns contextual information in the encoding step via either a feedforward network with context window or a bidirectional LSTM~\cite{hochreiter1997}.
Finally, in prediction step, the tagger predicts the POS tags from the output of the encoding step using either a softmax or a CRF layer.

\begin{figure}[!t]
\includegraphics[width=\linewidth]{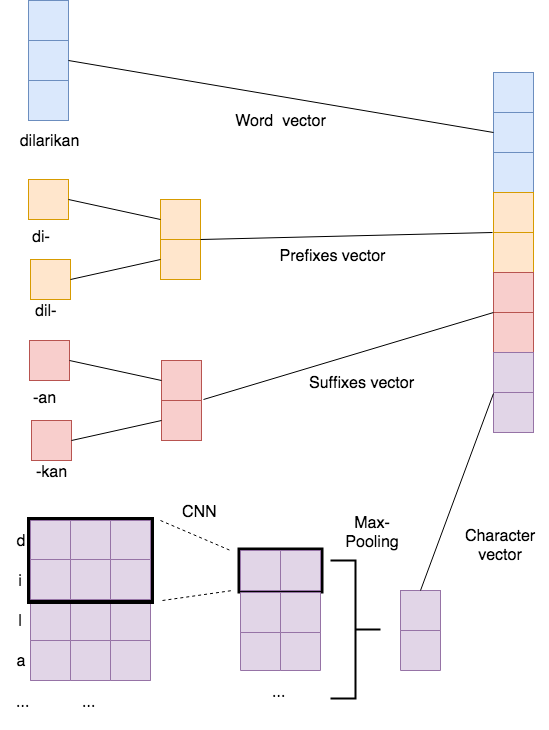}
\caption{Illustration of the embedding step. The word and its affixes are embedded to obtain their vector representations. Character embeddings of the word are composed with a max-pooled CNN. The final word embedding is the concatenation of all the result vectors.}\label{fig:embed}
\end{figure}

\textbf{Embedding}.~~In the embedding step, the tagger obtains vector representations of each word and additional features. We experimented with several additional features: prefixes, suffixes, and characters. Prefix features are the first 2 and 3 characters of the word. Likewise, suffix features are the last 2 and 3 characters of the word.\footnote{If the word has less than 3 characters, then all the prefixes and suffixes are equal to the word itself.} For the character features, we followed~\cite{ma2016} by embedding each character and composing the resulting vectors with a max-pooled CNN. The final embedding of a word is then the concatenation of all these vectors. Fig.~\ref{fig:embed} shows an illustration of the process.

\textbf{Encoding}.~~In the encoding step, the tagger learns contextual information by using either a feedforward network with context window or a bidirectional LSTM (biLSTM). The feedforward network accepts as input the concatenation of the  embedding of the current word and $d$ preceding and succeeding words for some context window size $d$. Formally, given a sequence of word embedding $\textbf{x}_1, \textbf{x}_2, \ldots, \textbf{x}_n$, the input of the feedforward network at timestep $t$ is
\begin{equation}
    \textbf{z}_t = \textbf{x}_{t-d} \oplus \ldots \oplus \textbf{x}_t \oplus \ldots \oplus \textbf{x}_{t+d}
\end{equation}
where $\oplus$ denotes a concatenation. The feedforward network then computes
\begin{align}
    \textbf{o}_t
    &= \text{FF}(\textbf{z}_t) \\
    &= W^{(2)} (\tanh(W^{(1)} \textbf{z}_t) \ast \textbf{r}_t)
\end{align}
where $\textbf{o}_t$ is the output vector, $\textbf{r}_t$ is a dropout mask vector, and $W^{(1)}, W^{(2)}$ are parameters.\footnote{There are bias vectors included as parameters, but we omit them from the equation for brevity.} The output vector $\textbf{o}_t$ has length equal to the number of possible tags. Its $j$-th component defines the (unnormalized) log probability of the $t$-th word having tag $j$.

On the other hand, the biLSTM accepts as input the sequence of word embeddings, and for each timestep, the output from the forward and backward LSTM are concatenated to form the final output. Formally, the output at each timestep $t$ can be expressed as
\begin{equation}
    \textbf{h}_t = \overrightarrow{\textbf{h}}_t \oplus \overleftarrow{\textbf{h}}_t
\end{equation}
where
\begin{align}
    \overrightarrow{\textbf{h}}_t
    &= \overrightarrow{\text{LSTM}}(\textbf{x}_t, \overrightarrow{\textbf{h}}_{t-1}) \\
    \overleftarrow{\textbf{h}}_t
    &= \overleftarrow{\text{LSTM}}(\textbf{x}_t, \overleftarrow{\textbf{h}}_{t-1})
\end{align}
The vector $\textbf{h}_t$ is then passed through $\text{FF}(\cdot)$ as before to obtain $\textbf{o}_t$.

\textbf{Prediction}.~~In the prediction step, the tagger predicts the POS tag of the $t$-th word based on the output vector $\textbf{o}_t$. We tested two approaches: a softmax layer with greedy decoding and a CRF layer with Viterbi decoding. With a softmax layer, the tagger simply normalizes $\textbf{o}_t$ and predicts using greedy decoding, i.e. picking the tag with the highest probability. In contrast, with a CRF layer, the tagger treats $\textbf{o}_t$ as emission probability scores, models the tag-to-tag transition probability scores, and uses Viterbi algorithm to select the most probable tag sequence as the prediction. We refer readers to~\cite{lample2016} to read more about how the CRF layer and Viterbi decoding work. We want to note that when we only embed words, encode using feedforward network, and predict using greedy decoding, the tagger is effectively the same as that in~\cite{abka2016}. Also, when only the word and character features are used, with a biLSTM and CRF layer, the tagger is effectively the same as that in~\cite{ma2016}. Our implementation code is available online.\footnote{https://github.com/kmkurn/id-pos-tagging}

\subsection{Experiments Setup}

For all models, we preprocessed the dataset by lowercasing all words, except when the characters were embedded. For the CRF model, we used L2 regularization whose coefficient was tuned to the development set. As we mentioned previously, we tuned the context window size $d$ to the development set as well.

For the neural tagger, we set the size of the word, affix, and character embedding to 100, 20, and 30 respectively. We applied dropout regularization to the embedding layers. The max-pooled CNN has 30 filters for each filter width. We set the feedforward network and the biLSTM to have 100 hidden units. We put a dropout layer before the biLSTM input layer. We tuned the learning rate, dropout rate, context window size, and CNN filter width to the development set. As we said earlier, we experimented with different configurations in the embedding, encoding, and prediction step. We evaluated each configuration on the development set as well.

At training time, we used a batch size of 8, decayed the learning rate by half if the $F_1$ score on the development set did not improve after 2 epochs, and stopped the training early if the score still did not improve after decaying the learning rate 5 times. To address the exploding gradient problem, we normalized the gradient norm at 1, following the suggestion in~\cite{reimers2017}. To handle the out-of-vocabulary problem, we converted singleton words and affixes occurring fewer than 5 times in the training data into a special token for unknown words/affixes.

\subsection{Evaluation}

Since the dataset is highly imbalanced (majority of words are nouns), using accuracy score as the evaluation metric is not appropriate as it gives a high  score to a model that always predicts nouns regardless of input. Therefore, we decided to use $F_1$ score which considers both precision and recall of the predictions.

Since there are multiple tags, there are two flavors to compute an overall $F_1$ score: micro and macro average. For POS tagging task where the tags do not span multiple words, micro-average $F_1$ score is exactly the same as accuracy score. Thus, macro-average $F_1$ score is our only option. However, there is still an issue. Macro-average $F_1$ score computes the overall $F_1$ score by averaging the $F_1$ score of each tag. This approach means that when the model wrongly predicts a rarely occurring tag (e.g., foreign word), it is penalized as heavily as it does a frequent tag. To address this problem, we used weighted macro-average $F_1$ score which takes into account the tag proportion imbalance. It computes the weighted average of the scores where each weight is equal to the corresponding tag's proportion in the dataset. This functionality is available in the \texttt{scikit-learn} library.\footnote{http://scikit-learn.org/stable/modules/generated/sklearn.metrics.f1\_score.html}

\section{Results and Discussion}

\begin{table}[!t]
    \renewcommand{\arraystretch}{1.3}
    \caption{Dev $F_1$ score of each neural tagger architecture}\label{tbl:dev-neural}
    \centering
    \begin{tabular}{|l|r|}
        \hline
        \multicolumn{1}{|c}{\textbf{Architecture}} & \multicolumn{1}{|c|}{\textbf{Mean} $\mathbf{F_1}$} \\
        \hline
        Feedforward + softmax & 97.25 \\
        Feedforward + CRF & 97.46 \\
        biLSTM + softmax & 97.57 \\
        biLSTM + CRF & \textbf{97.60} \\
        \hline
    \end{tabular}
\end{table}

\begin{table}[!t]
    \renewcommand{\arraystretch}{1.3}
    \caption{Test $F_1$ score of each method, averaged over 5 folds}\label{tbl:results}
    \centering
    \begin{tabular}{|l|r|}
        \hline
        \multicolumn{1}{|c}{\textbf{Method}} & \multicolumn{1}{|c|}{$\mathbf{F_1}$} \\
        \hline
        \textsc{Major} & 9.39 (0.21) \\
        \textsc{Memo} & 90.62 (0.82) \\
        Rashel et al.~\cite{rashel2014building} & 85.77 (0.22)\\
        CRF & 96.22 (0.22) \\
        biLSTM + CRF & \textbf{97.47 (0.11)} \\
        \hline
    \end{tabular}
\end{table}

\begin{table}[!t]
    \renewcommand{\arraystretch}{1.3}
    \caption{Feature ablation of the best neural tagger}\label{tbl:ablation}
    \centering
    \begin{tabular}{|l|r|}
    \hline
    \multicolumn{1}{|c}{\textbf{Neural tagger}} & \multicolumn{1}{|c|}{$\mathbf{F_1}$} \\
    \hline
    biLSTM + CRF & 96.06 (+0.00) \\
    biLSTM + CRF + chars & 97.42 (+1.36) \\
    biLSTM + CRF + chars + prefix & 97.50 (+0.08) \\
    biLSTM + CRF + chars + prefix + suffix & 97.60 (+0.10) \\
    \hline
    \end{tabular}
\end{table}

Firstly, we report on our tuning experiments for the neural tagger. Table~\ref{tbl:dev-neural} shows the evaluation results of the many configurations of our neural tagger on the development set. We group the results by the encoding and prediction step configuration. For each group, we show the highest $F_1$ score among many embedding configurations. As we can see, biLSTM with CRF layer achieves 97.60 $F_1$ score, the best score on the development set. This result agrees with many previous work in neural sequence labeling that a bidirectional LSTM with CRF layer performs best~\cite{rei2016,lample2016,ma2016}. Therefore, we will use this tagger to represent the neural model hereinafter.

\begin{figure}[!t]
\includegraphics[width=\linewidth]{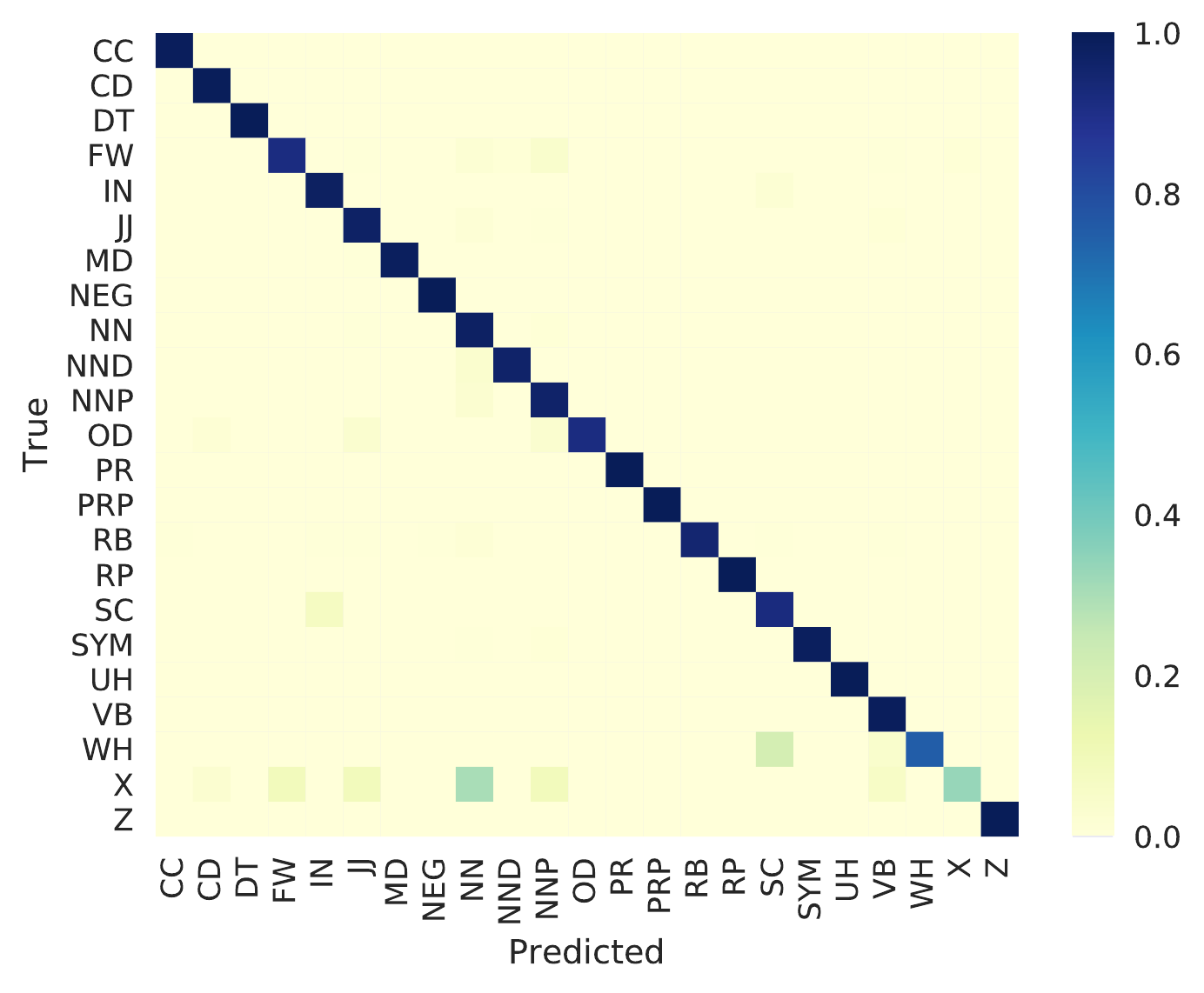}
\caption{Confusion matrix of the best biLSTM with CRF tagger from the development set of the first fold. The tagger seems to have difficulties dealing with words annotated as X and confuse WH as SC.}\label{fig:cf}
\end{figure}

To understand the performance of the neural model for each tag, we plot the confusion matrix from the development set of the first fold in Fig.~\ref{fig:cf}. The figure shows that the model can predict most tags almost perfectly, except for X and WH tag. The X tag is described as "a word or part of a sentence which its category is unknown or uncertain".\footnote{http://bahasa.cs.ui.ac.id/postag/downloads/Tagset.pdf} The X tag is rather rare, as it only appears 397 times out of over 250K tokens. Some words annotated as X are typos and slang words. Some foreign terms and abbreviations are also annotated with X. The model might get confused as such words are usually tagged with a noun tag (NN or NNP). We also see that the model seems to confuse question words (WH) such as \textit{apa} (what) or \textit{siapa} (who) as SC since these words may be used in subordinate clauses as well. Looking at the data closely, we found that the tagging of such words are inconsistent. This inconsistency contributes to the inability of the model to distinguish the two tags well.

Next, we present the result of evaluating the baselines and other comparisons on the test set in Table~\ref{tbl:results}. The $F_1$ scores are averaged over the 5 cross-validation folds. We see that \textsc{Major} baseline performs very poorly compared to the \textsc{Memo} baseline, which surprisingly achieves over 90 $F_1$ points. This result suggests that \textsc{Memo} is a more suitable baseline for this dataset in contrast with \textsc{Major}. The result also provides evidence to the usefulness of our evaluation metric which heavily penalizes a simple majority vote model. Furthermore, we notice that the rule-based tagger by Rashel et al.~\cite{rashel2014} performs worse than \textsc{Memo}, indicating that \textsc{Memo} is not just suitable but also quite a strong baseline. Moving on, we observe how CRF has 6 points advantage over \textsc{Memo}, signaling that incorporating contextual features and modeling tag-to-tag transitions are useful. Lastly, the biLSTM with CRF tagger performs the best with 97.47 $F_1$ score.

To understand how each feature in the embedding step affects the neural tagger, we performed feature ablation on the development set and put the result in Table~\ref{tbl:ablation}. We see that with only words as features (first row), the neural tagger only achieves 96.06 $F_1$ score. Employing character features boosts the score up to 97.42, a gain of 1.36 points. Adding prefix and suffix features improves the performance further by 0.08 and 0.10 points respectively. From this result, we see that it is the character features that positively affect the neural tagger the most.

\section{Conclusion}

We experimented with several baselines and comparisons for Indonesian POS tagging task. Our comparisons include a rule-based tagger, a well-established probabilistic model for sequence labeling (CRF), and a neural model. We tested many configurations for our neural model: the features (words, affixes, characters), the architecture (feedforward, biLSTM), and the output layer (softmax, CRF). We evaluated all our models on the IDN Tagged Corpus~\cite{dinakaramani2014}, a manually annotated and publicly available Indonesian POS tagging dataset. Our best model achieves 97.47 $F_1$ score, a new state-of-the-art result on the dataset. We make our cross-validation split available publicly to serve as a benchmark for future work.





%
\bibliographystyle{IEEEtranBST/IEEEtran}
\bibliography{pos-tagging}

\begin{thebibliography}{10}
\providecommand{\url}[1]{#1}
\csname url@samestyle\endcsname
\providecommand{\newblock}{\relax}
\providecommand{\bibinfo}[2]{#2}
\providecommand{\BIBentrySTDinterwordspacing}{\spaceskip=0pt\relax}
\providecommand{\BIBentryALTinterwordstretchfactor}{4}
\providecommand{\BIBentryALTinterwordspacing}{\spaceskip=\fontdimen2\font plus
\BIBentryALTinterwordstretchfactor\fontdimen3\font minus
  \fontdimen4\font\relax}
\providecommand{\BIBforeignlanguage}[2]{{%
\expandafter\ifx\csname l@#1\endcsname\relax
\typeout{** WARNING: IEEEtran.bst: No hyphenation pattern has been}%
\typeout{** loaded for the language `#1'. Using the pattern for}%
\typeout{** the default language instead.}%
\else
\language=\csname l@#1\endcsname
\fi
#2}}
\providecommand{\BIBdecl}{\relax}
\BIBdecl

\bibitem{luo2015joint}
G.~Luo, X.~Huang, C.-Y. Lin, and Z.~Nie, ``Joint entity recognition and
  disambiguation,'' in \emph{Proceedings of the 2015 Conference on Empirical
  Methods in Natural Language Processing}, 2015, pp. 879--888.

\bibitem{aguilar2017multi}
G.~Aguilar, S.~Maharjan, A.~P.~L. Monroy, and T.~Solorio, ``A multi-task
  approach for named entity recognition in social media data,'' in
  \emph{Proceedings of the 3rd Workshop on Noisy User-generated Text}, 2017,
  pp. 148--153.

\bibitem{sennrich-haddow:2016:WMT}
\BIBentryALTinterwordspacing
R.~Sennrich and B.~Haddow, ``{Linguistic Input Features Improve Neural Machine
  Translation},'' in \emph{{Proceedings of the First Conference on Machine
  Translation}}.\hskip 1em plus 0.5em minus 0.4em\relax Berlin, Germany:
  Association for Computational Linguistics, August 2016, pp. 83--91. [Online].
  Available: \url{http://www.aclweb.org/anthology/W16-2209.pdf}
\BIBentrySTDinterwordspacing

\bibitem{niehues2017exploiting}
J.~Niehues and E.~Cho, ``Exploiting linguistic resources for neural machine
  translation using multi-task learning,'' \emph{arXiv preprint
  arXiv:1708.00993}, 2017.

\bibitem{dyer2016}
C.~Dyer, A.~Kuncoro, M.~Ballesteros, and N.~A. Smith, ``Recurrent neural
  network grammars,'' in \emph{Proceedings of the 2016 {{Conference}} of the
  {{North American Chapter}} of the {{Association}} for {{Computational
  Linguistics}}: {{Human Language Technologies}}}.\hskip 1em plus 0.5em minus
  0.4em\relax San Diego, California: {Association for Computational
  Linguistics}, 2016, pp. 199--209.

\bibitem{pisceldo2009}
F.~Pisceldo, M.~Adriani, and R.~Manurung,
  ``\BIBforeignlanguage{en}{Probabilistic {{Part Of Speech Tagging}} for
  {{Bahasa Indonesia}}},'' 2009, p.~6.

\bibitem{wicaksono2010}
A.~F. Wicaksono and A.~Purwarianti, ``{{HMM Based Part}}-of-{{Speech Tagger}}
  for {{Bahasa Indonesia}},'' Jan. 2010.

\bibitem{rashel2014}
F.~Rashel, A.~Luthfi, A.~Dinakaramani, and R.~Manurung, ``Building an
  {{Indonesian}} rule-based part-of-speech tagger,'' in \emph{2014
  {{International Conference}} on {{Asian Language Processing}} ({{IALP}})},
  Oct. 2014, pp. 70--73.

\bibitem{abka2016}
A.~F. Abka, ``Evaluating the use of word embeddings for part-of-speech tagging
  in {{Bahasa Indonesia}},'' in \emph{2016 {{International Conference}} on
  {{Computer}}, {{Control}}, {{Informatics}} and Its {{Applications}}
  ({{IC3INA}})}, Oct. 2016, pp. 209--214.

\bibitem{ma2016}
X.~Ma and E.~Hovy, ``\BIBforeignlanguage{en}{End-to-end {{Sequence Labeling}}
  via {{Bi}}-directional {{LSTM}}-{{CNNs}}-{{CRF}}},'' in
  \emph{\BIBforeignlanguage{en}{Proceedings of the 54th Annual Meeting of the
  Association for Computational Linguistics (Volume 1: Long Papers)}}.\hskip
  1em plus 0.5em minus 0.4em\relax {Association for Computational Linguistics},
  2016, pp. 1064--1074.

\bibitem{rei2016}
M.~Rei, G.~K.~O. Crichton, and S.~Pyysalo, ``Attending to characters in neural
  sequence labeling models,'' in \emph{Proceedings of {{COLING}} 2016, the 26th
  {{International Conference}} on {{Computational Linguistics}}: {{Technical
  Papers}}}, Osaka, Japan, 2016, pp. 309--318.

\bibitem{dinakaramani2014}
A.~Dinakaramani, F.~Rashel, A.~Luthfi, and R.~Manurung,
  ``\BIBforeignlanguage{en}{Designing an {{Indonesian}} part of speech tagset
  and manually tagged {{Indonesian}} corpus},'' in
  \emph{\BIBforeignlanguage{en}{2014 {{International Conference}} on {{Asian
  Language Processing}} ({{IALP}})}}.\hskip 1em plus 0.5em minus 0.4em\relax
  {IEEE}, 2014, pp. 66--69.

\bibitem{lafferty2001}
J.~D. Lafferty, A.~McCallum, and F.~C.~N. Pereira, ``Conditional {{Random
  Fields}}: {{Probabilistic Models}} for {{Segmenting}} and {{Labeling Sequence
  Data}},'' in \emph{Proceedings of the {{Eighteenth International Conference}}
  on {{Machine Learning}}}, ser. ICML '01.\hskip 1em plus 0.5em minus
  0.4em\relax San Francisco, CA, USA: {Morgan Kaufmann Publishers Inc.}, 2001,
  pp. 282--289.

\bibitem{collobert2011}
R.~Collobert, J.~Weston, L.~Bottou, M.~Karlen, K.~Kavukcuoglu, and P.~Kuksa,
  ``Natural language processing (almost) from scratch,'' \emph{Journal of
  Machine Learning Research}, vol.~12, no. Aug, pp. 2493--2537, 2011.

\bibitem{rashel2014building}
F.~Rashel, A.~Luthfi, A.~Dinakaramani, and R.~Manurung, ``Building an
  indonesian rule-based part-of-speech tagger,'' in \emph{Asian Language
  Processing (IALP), 2014 International Conference on}.\hskip 1em plus 0.5em
  minus 0.4em\relax IEEE, 2014, pp. 70--73.

\bibitem{larasati2011indonesian}
S.~Larasati, V.~Kubo{\v{n}}, and D.~Zeman, ``Indonesian morphology tool
  (morphind): Towards an indonesian corpus,'' \emph{Systems and Frameworks for
  Computational Morphology}, pp. 119--129, 2011.

\bibitem{hochreiter1997}
S.~Hochreiter and J.~Schmidhuber, ``Long short-term memory,'' \emph{Neural
  Computation}, vol.~9, no.~8, p. 1735–1780, Nov 1997.

\bibitem{lample2016}
\BIBentryALTinterwordspacing
G.~Lample, M.~Ballesteros, S.~Subramanian, K.~Kawakami, and C.~Dyer, ``Neural
  architectures for named entity recognition,'' in \emph{Proceedings of the
  2016 Conference of the North American Chapter of the Association for
  Computational Linguistics: Human Language Technologies}.\hskip 1em plus 0.5em
  minus 0.4em\relax Association for Computational Linguistics, 2016, p.
  260–270. [Online]. Available:
  \url{http://aclanthology.coli.uni-saarland.de/pdf/N/N16/N16-1030.pdf}
\BIBentrySTDinterwordspacing

\bibitem{reimers2017}
\BIBentryALTinterwordspacing
N.~Reimers and I.~Gurevych, ``Reporting score distributions makes a difference:
  Performance study of lstm-networks for sequence tagging,'' in
  \emph{Proceedings of the 2017 Conference on Empirical Methods in Natural
  Language Processing}.\hskip 1em plus 0.5em minus 0.4em\relax Association for
  Computational Linguistics, Sep 2017, p. 338–348. [Online]. Available:
  \url{https://www.aclweb.org/anthology/D17-1035}
\BIBentrySTDinterwordspacing

\end{thebibliography}

\end{document}